%% file: main.tex
\documentclass[10pt,letterpaper]{article}

\usepackage{cogsci}

\cogscifinalcopy %%% Uncomment this line for the final submission

\usepackage[
  style=apa,
  natbib=true,
  annotation=false,
  hyperref,
]{biblatex}
\usepackage{float} %%% Roger Levy added this and changed figure/table placement to [H] for conformity to Word template, though floating tables and figures to top is still generally recommended!

\author[1]{\mbox{Nathaniel Imel (ni2128@nyu.edu)}}
\author[1]{\mbox{Noga Zaslavsky (nogaz@nyu.edu)}}
\affil[1]{Department of Psychology, New York University}

\input{header_modified.tex} % I needed to remove the algorithmic package, it was breaking alg line numbers

\usepackage{appendix}

\theoremstyle{definition}
\newtheorem{definition}{Definition}

\newtheorem{theorem}{Theorem}

\usepackage{graphicx} % images

\usepackage{booktabs} % tables
\usepackage{multirow}
\usepackage[table]{xcolor}

\usepackage{times}
\usepackage{hyperref}
\usepackage{cleveref}
\usepackage{balance}

\usepackage{algorithm}
\usepackage{algpseudocode}

\newcommand\blfootnote[1]{%
  \begingroup
  \renewcommand\thefootnote{}\footnote{#1}%
  \addtocounter{footnote}{-1}%
  \endgroup
}

\newcommand{\SM}[1]{\href{#1}{SM}}
\newcommand{\SMurlFigures}{https://tinyurl.com/sm-figures}
\newcommand{\SMWebApp}{\href{\SMurlFigures}{SM interactive app}}
\newcommand{\apndx}{Appendix}

\title{On convexity and efficiency in semantic systems}

\begin{document}

\maketitle

\begin{abstract}
    There are two widely held characterizations of human semantic category systems: (1)~they form convex partitions of conceptual spaces, and (2)~they are efficient for communication. While prior work observed that convexity and efficiency co-occur in color naming, the analytical relation between them and why they co-occur have not been well understood. We address this gap by combining analytical and empirical analyses that build on the Information Bottleneck (IB) framework for semantic efficiency. First, we show that convexity and efficiency are distinct in the sense that neither entails the other: there are convex systems which are inefficient, and optimally-efficient systems that are non-convex. Crucially, however, the IB-optimal systems are mostly convex in the domain of color naming, explaining the main empirical basis for the convexity approach. Second, we show that efficiency is a stronger predictor for discriminating attested color naming systems from hypothetical variants, with convexity adding negligible improvement on top of that. Finally, we discuss a range of empirical phenomena that convexity cannot account for but efficiency can. Taken together, our work suggests that while convexity and efficiency can yield similar structural observations, they are fundamentally distinct, with efficiency providing a more comprehensive account of semantic typology.
    \blfootnote{Supplementary materials (SM) for this paper are available at \url{\SMurlFigures}.}

 
\end{abstract}

\section{Introduction}

Across cultures and domains of human experience, semantic categories exhibit constrained variation. In domains such as color, kinship, artifacts, and space, only a small subset of logically possible category systems appear to be attested. A central question in the study of semantic typology is therefore: what forces shape the structure of these category systems? 

\begin{figure*}[t!]
    \centering
    \includegraphics[
    width=.95\linewidth,
    % trim = <left> <bottom> <right> <top>
    trim=0in 2.5in 0in 0in,
    clip
    ]{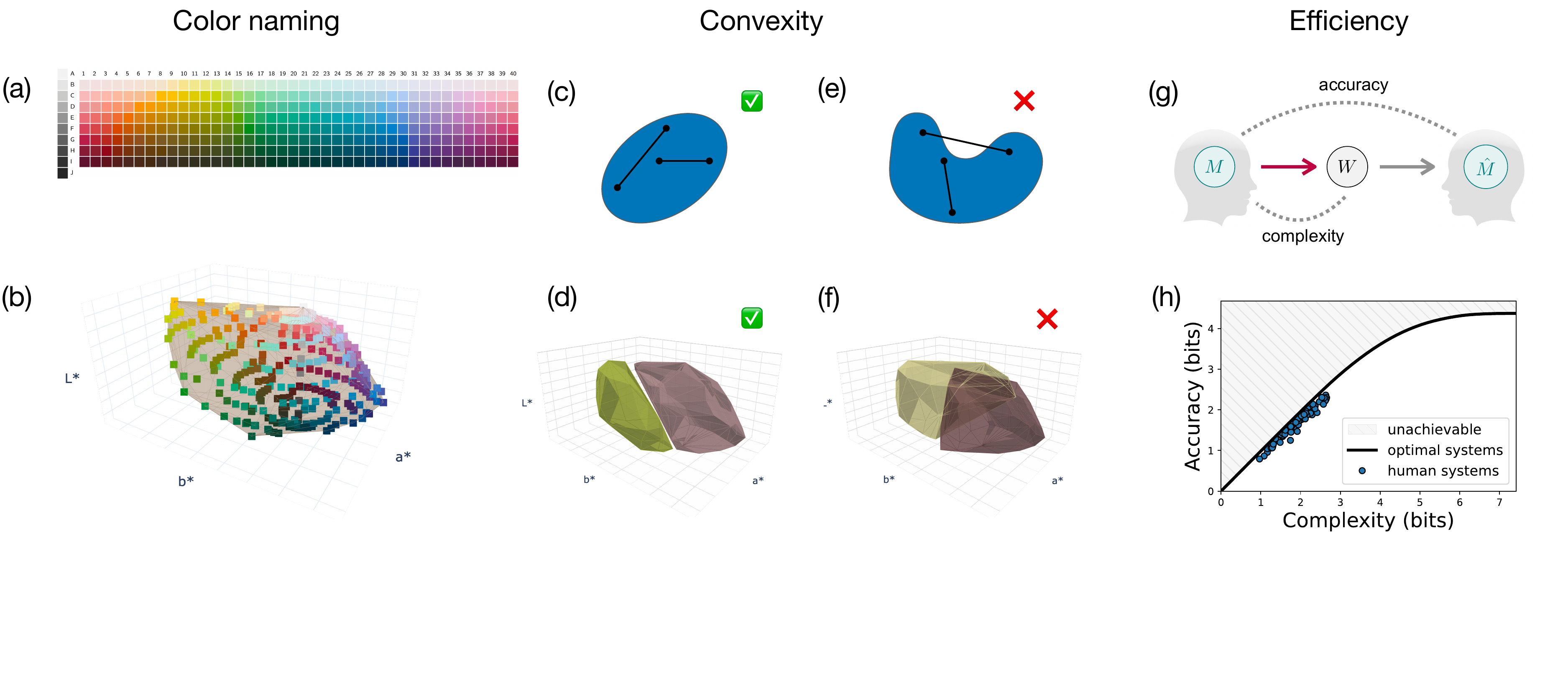}
    \caption{
    \textbf{(a)} The WCS color naming grid, standard in color naming research.
    \textbf{(b)} The WCS color stimuli visualized in the 3D CIELAB perceptual color space (see our \SMWebApp\ for exploring the structure of this 3D space).
    \textbf{(c-e)} Illustration of convex and non-convex sets. 
    \textbf{(d-f)} Examples of convex and non-convex color category systems in CIELAB.
    \textbf{(g)}~The IB communication model (see main text).
    \textbf{(h)}~The IB theoretical bound (black curve) and empirical results (blue dots) in the domain of color naming, reproduced from \citet{Zaslavsky2018Efficient}. 
    }
    \label{fig:stagesetting}
\end{figure*}

Two influential approaches to this question have been proposed in the literature, one based on the notion of convexity and one based on the notion of communicative efficiency~(\Cref{fig:stagesetting}).
The convexity approach~\citep{gardenfors1990induction, gardenfors2000conceptual, gardenfors2014geometry} argues that naturally occurring semantic categories correspond to convex regions in an underlying conceptual space. Intuitively, a category is convex if for any two elements it contains, the category also contains all points in between them in the conceptual space. This approach has been supported empirically primarily in the domain of color naming~\citep{gardenfors2000conceptual,Jager2010Natural}, as color is a continuous domain with a well-established perceptual space, leading to a natural quantitatively-testable instantiation of the notion of convexity. Beyond the domain of color, \citet{Douven2016Vagueness} show that in a synthetic space of containers~\citep[inspired from ][]{Labov1973Boundaries}, sets of objects people identified as `typical' for categories formed mostly convex regions of a space obtained via multidimensional scaling. In the domain of modality, \citet{Steinert-Threlkeld2023Semantic} propose a semantic universal equivalent to convexity, and \citet{Chemla2019Connecting} explore how {connectedness}---a property weaker than convexity but still central to G\"ardenfors' theory---may extend to both content and function words.

The efficient communication approach argues languages balance two competing pressures: maximizing communicative accuracy, or informativeness, and minimizing communicative effort, or cost~\citep{zipf1949least,Rosch1978Principles,Kemp2018Semantic,Gibson2019}. While this notion of efficiency has been formulated in the literature in various ways, here we build on the Information Bottleneck (IB) formulation~\citep{Zaslavsky2018Efficient}, which is based on a formal information-theoretic complexity-accuracy optimality principle~\citep{Tishby1999Information} that is grounded in rate-distortion theory~\citep{Shannon1959Coding, Berger1971Rate}, the mathematical theory of lossy data compression. This framework has been gaining broad cross-linguistic support across semantic domains, including color~\citep{Zaslavsky2018Efficient,Zaslavsky2022Evolution}, visual objects~\citep{Zaslavsky2019Semantic,taliaferro2025bilinguals}, personal pronouns~\citep{Zaslavsky2021Lets}, and locomotion~\citep{langlois2025}, yielding a formal predictive characterization of semantic systems in terms of the IB principle.

While convexity and efficiency have been derived independently, recent work has started to explore the relationship between them. In particular, it has been noted that convexity and efficiency coincide in the domain of color naming~\citep{Douven2020What,Gardenfors2024Natural,Koshevoy2025Convexity,Skinner2025Convexity, Bruneau-Bongard2025Assessing}, leading to a conjecture that convexity may be ``instrumental in facilitating efficient communication of concepts''~\citep[][p.880]{Gardenfors2024Natural}.  This conjecture, however, has not been proven, and more broadly, the precise mathematical and empirical relationship between convexity and efficiency has not been well-understood.

Here, we address this gap in the literature by combining mathematical proofs and empirical analyses that clarify the relationship between convexity and efficiency. First, we show that \citet{Gardenfors2024Natural}'s conjecture does not hold in general. That is, we show that convexity and efficiency are distinct and neither necessitates the other: convexity can lead to inefficiency, and optimally-efficient systems are not necessarily convex. Crucially, however, the IB-optimal systems are mostly convex in the domain of color naming, explaining the main empirical basis for the convexity approach purely in terms of efficiency. Second, we show that efficiency is a stronger predictor for discriminating between attested color naming systems and hypothetical variants, with convexity adding negligible predictive value to efficiency. Finally, we discuss a range of empirical phenomena that convexity cannot account for but efficiency can, including adaptation to environmental factors and language change. Taken together, our work suggests that while convexity and efficiency can yield similar structural observations, they are fundamentally distinct, with efficiency providing a more comprehensive account of semantic typology.

\section{Theoretical Background}

We begin with an overview of the theoretical foundations for our work, formally defining convexity and efficiency in sematic systems. We use the following definitions and notation: a category system is defined by a mapping $P: \mathcal{U}\rightarrow\mathcal{W}$, where $\mathcal{U}$ is domain consisting of a set of objects in the world, and $\mathcal{W}$ is a set of words that can be used to describe these objects. For each $w\in\mathcal{W}$, we define its category extension, $C_P(w)$, by the set of objects that are mapped to it, i.e., $C_P(w) = \{u\in \mathcal{U}: P(u)=w\}$. Denote by $k=|\mathcal{W}|$ the number of categories in the system. For $\cU\in\reals^d$, the convex-hull of a set $C$ is defined by
$\mathrm{co}(C) = \{z=\alpha x + (1-\alpha) y : x,y\in C, \alpha\in[0,1]\}$.

\subsection{Convexity}
The notion that semantic categories might correspond to convex regions in an underlying semantic space is associated most prominently with the work of G{\"a}rdenfors \citep{gardenfors1990induction, gardenfors2000conceptual, gardenfors2014geometry}. The most relevant claim in this literature for our work is that convexity may be a necessary structural constraint on natural properties. \citet{gardenfors2000conceptual} posits: ``most properties expressed by simple words in natural languages can be analyzed as natural properties'' (p. 76), where ``a natural property is a convex region of a domain in a conceptual space'' (p. 71).
The formulation of convexity rests on a shared structural intuition: if two points belong to a category, any point in between them should also belong to that category. Identifying an appropriate ``in-betweenness'' relation, $B$, is therefore critical for formalizing the convexity of categories \citep{gardenfors2000conceptual}. Such a relation should be defined over a domain $\mathcal{U}$, such that for any $x,y,z\in\mathcal{U}$, if $B(x,z,y)$ is True then $z$ is in-between $x$ and $y$. Given $B$, a category $C$ is convex if
\begin{equation}
    \label{convexity_betweenness}
    \forall x, y \in C, \forall z\in\mathcal{U}: B(x, z, y) \implies z \in C.
\end{equation}
In continuous domains, such as color, in-betweenness can be defined by the shortest line segment connecting two points. This notion of convexity generalizes the standard definition of a convex set over continuous vector spaces, in which $C$ is convex if for any $x,y\in C$, the linear combination of these points $z=\alpha x + (1-\alpha) y$, for $\alpha \in [0,1]$, is also contained in $C$. This generalization is crucial because it can also be applied in discrete domains~\citep{Chemla2019Connecting}, which is necessary for characterizing semantic categories across the lexicon.

Testing for convexity in practice requires special considerations. Empirical data can be noisy, yielding estimated categories that may not be perfectly convex even if the underlying categories are convex. Furthermore, language evolution is a complex process that may yield categories which are nearly convex but not perfectly so. Therefore, graded notions of convexity have been considered in the literature.

In the domain of color, for example, where $\cU$ corresponds to colors in CIELAB~(\Cref{fig:stagesetting}b), a perceptual space embedded in $\reals^3$, \citet{Steinert-Threlkeld2020Ease} proposed to estimate the degree of convexity for a category by the ratio of its extension to the size of its convex hull, i.e.,
\eqn{
0\le\frac{|C_P(w)|}{|\text{co}(C_P(w))|}\le1\,,
}
such that a perfectly convex category achieves a value of $1$. The degree of convexity for a system $P$ is then the weighted average over its categories, i.e.,
\eqn{
    \label{eq:convexity_consistency}
    \mathrm{Conv}[P] = \frac{\sum_w |C_P(w)| \frac{|C_P(w)|}{|\text{co}(C_P(w))|}}{\sum_w |C_P(w)|}.
}
Using \Cref{eq:convexity_consistency}, \citet{Koshevoy2025Convexity} and \citet{Bruneau-Bongard2025Assessing} find that systems of color categories across the languages of the World Color Survey~\citep{berlin1969,Kay2009World} show high degrees of convexity, resonating with earlier findings by~\citet{Jager2010Natural}. An example of such an approximately convex system, and the convex hulls of its categories, can be seen in \Cref{fig:inefficient}c.

\subsection{Efficiency}

The notion of efficient communication is based on the intuition that languages balance informativeness, which supports successful communication, against cognitive cost~\citep[see][for reviews]{Kemp2018Semantic, Gibson2019}. To formulate this notion, we use the theoretical framework of~\citet{Zaslavsky2018Efficient} which is based on the Information Bottleneck (IB) principle~\citep{Tishby1999Information}.
According to this framework, languages evolve under pressure to efficiently compress meanings into words, or categories, by near-optimally satisfying the IB tradeoff between the informational complexity, or compressibility, and communicative accuracy of the lexicon. 
The definition of this optimality principle is based on the following communication model (\Cref{fig:stagesetting}g): a speaker needs to communicate a target referent $u_t\in\cU$, sampled from a communicative need distribution $p(u_t)$. The speaker forms a mental representation, or belief state, $m_t\in\mathcal{M}$, defined as a probability distribution over $\mathcal{U}$, which captures the uncertainty (or belief) that the speaker has over the target. The speaker then maps its belief state to a word $w\in\mathcal{W}$ via an encoder $q(w|m)$, which can be probabilistic. A listener receives $w$ and constructs an estimator $\hat{m}_w$ of the speaker's belief state.

In the case of color, for example, the domain $\mathcal{U}$ is given by the set of target colors shown in \Cref{fig:stagesetting}a. To account for perceptual noise, each $m_t$ is modeled as a Gaussian over the perceptual CIELAB color space (\Cref{fig:stagesetting}b) centered at a corresponding target color. That is, the speaker represents each color as a Gaussian distribution and the listener's goal is to reconstruct the speaker's belief state over colors.

In order to communicate efficiently, the speaker and listener must jointly optimize the IB tradeoff between minimizing the complexity of their semantic system and maximizing its accuracy. Complexity in IB corresponds to the number of bits required for communication, formally captured by $I_q(M;W)$, the mutual information between the speaker's meanings and words. 
Accuracy is defined by $I_q(W;U)$, the information that the speaker's words maintain about the target world state, which is inversely related to the Kullback-Leibler (KL) divergence between the speaker's mental state and the listener's inferred state, $\mathbb{E}_q[D[M\|\hat{M}]]$~\citep[see][]{Harremoes2007Information,Zaslavsky2020InformationTheoretic}. An optimal semantic system is one that minimizes the IB objective function
\begin{equation}
    \label{eq:ib}
    \mathcal{F}_{\beta}[q] = I_q(M;W) - \beta I_q(W;U),
\end{equation}
where $\beta \geq 0$ controls the complexity-accuracy tradeoff. The solutions to this optimization problem define the IB theoretical limit of efficiency (\Cref{fig:stagesetting}h), where $\beta=0$ yields minimally complex and non-informative systems and as $\beta\rightarrow\infty$ the optimal systems become more complex while attaining the maximal possible accuracy given their level of complexity.

\begin{figure*}[t!]
    \centering
    \includegraphics[
    width=0.9\linewidth,
    % trim = <left> <bottom> <right> <top>
    trim = 0 2.35in 0 0,
    clip    
    ]{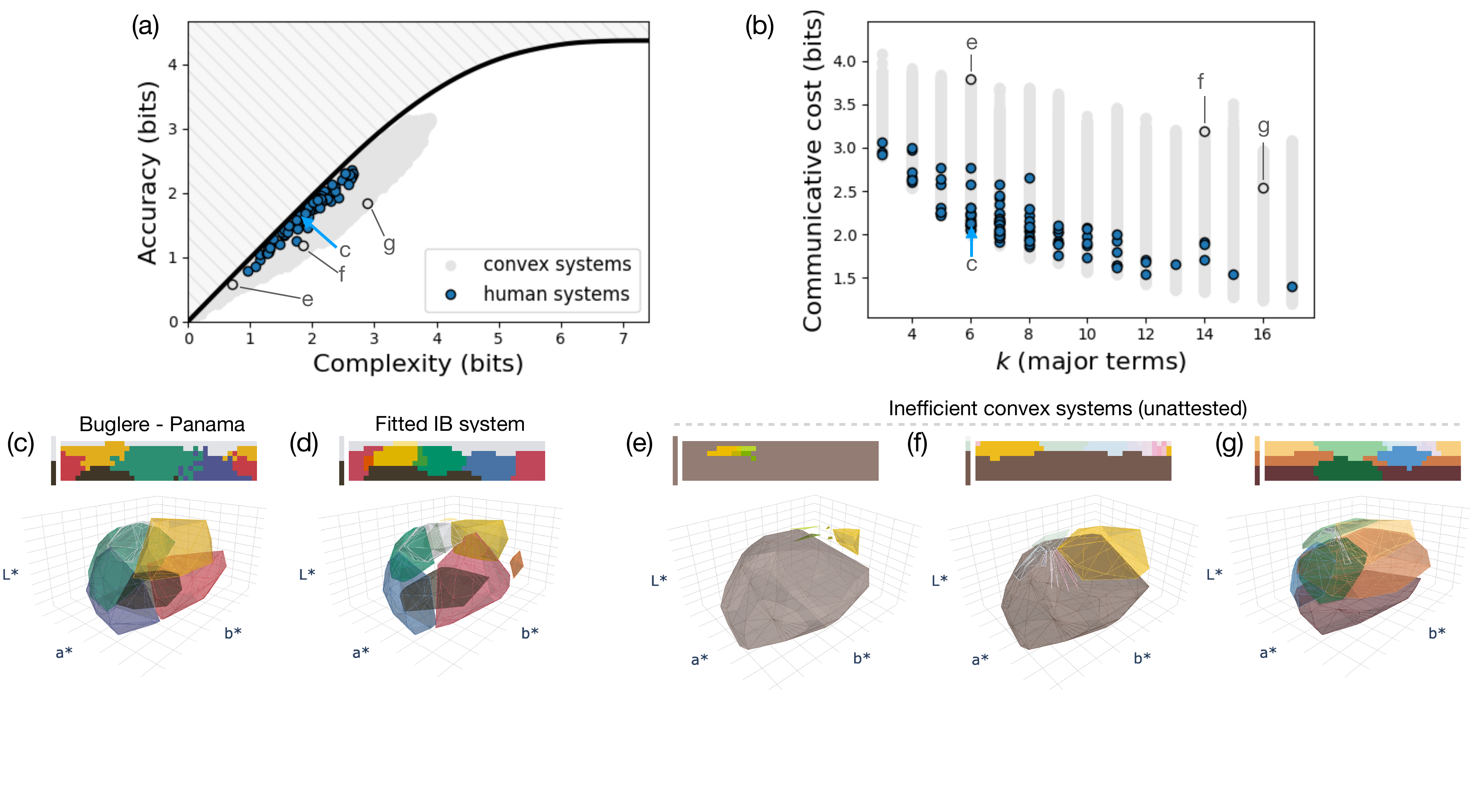}
    \caption{
    \textbf{(a)} IB trade-off for human color naming systems (blue circles) from the WCS and a large sample of artificial convex systems (gray circles). Black circles indicate costly convex systems, visualized in (e-g). \textbf{(b)} Same systems in (a), with $k$ (number of major terms) vs. communicative cost.
    \textbf{(c,d)} Example human color naming system (Buglere, Panama) and its fitted IB system.
    \textbf{(e-g)} Three example convex, inefficient category systems as partitions in the 2D WCS grid (top row) and convex hulls in the 3D CIELAB color space (bottom row). Interactive versions of these figures and additional systems can be viewed in our \SMWebApp.
    }
    \label{fig:inefficient}
\end{figure*}

To measure the efficiency of an encoder, or a semantic system,  \citet{Zaslavsky2018Efficient} defined the following measure of deviation from optimality:
\begin{equation}
    \label{eq:epsilon}
    \varepsilon[q] = \min_{\beta} \frac1\beta \left(\mathcal{F}_{\beta}[q] -  \mathcal{F}_{\beta}^*\right),
\end{equation}
where $\mathcal{F}_{\beta}^*$ is the optimal value of $\mathcal{F}_{\beta}$ (see \citealp{Zaslavsky2018Efficient} for more details). They predicted that languages are pressured to minimize this deviation from optimality, and that optimal IB systems should characterize attested systems' category structure (see \Cref{fig:inefficient}d for how such an example IB system fits an actual human color naming system).  This prediction has been initially supported in the case of color naming~\citep[\Cref{fig:stagesetting}h;][]{Zaslavsky2018Efficient,Zaslavsky2022Evolution} and has since been gaining broader cross-linguistic support across other semantic domains~\citep{ Zaslavsky2019Semantic,Zaslavsky2021Lets,Mollica2021Forms,langlois2025}.

\section{Revisiting G{\"a}rdenfors' conjecture}

\citet{Gardenfors2024Natural}'s conjecture that convexity plays an important role in facilitating efficiency has been motivated by two observations: First, empirically, convexity and efficiency appear to coincide in the domain of color naming. That is, as reviewed above, systems of color categories across languages tend to be both convex and efficient. Second, optimally-efficient color naming systems tend to be convex as well. \citet{Douven2020What} showed that an earlier notion of efficiency that is not based on the IB principle but rather on well-formedness~\citep{Regier2007Color}, yields convex partitions of color space. More recent studies~\citep{Koshevoy2025Convexity, Skinner2025Convexity} have extended this observation to IB.

To confirm these observations, we repeated a similar analysis by evaluating the degree of convexity in human and IB-optimal color category systems. The human systems were estimated from the World Color Survey~\citep[WCS;][]{berlin1969,Kay2009World}, which provides color naming data from 110 languages of non-industrialized societies. For each language $l$, we estimated a color category system $\hat{P}_l$ by mapping each color to the category induced by the modal term that was used to describe it. The IB systems were derived from the previously-established IB color naming model~\citep{Zaslavsky2018Efficient}.
Similar to the human data, for an IB-optimal encoder $q_{\beta}(w|m)$, we define a category system $P_{\beta}$ by mapping each color $c$ to its modal category $w_c = \argmax_w q_{\beta}(w|m_c)$, where $m_c$ is the speaker's mental representation of $c$.

We adopt~\Cref{eq:convexity_consistency} as a measure of convexity. We note, however, that for any empirically-estimated category system $\hat{P}$, the criterion $\mathrm{Conv}[\hat{P}]=1$ is a necessary but not sufficient condition for the true underlying $P$, from which the empirical data was generated, to be convex. We therefore argue that this measure does not capture the degree of convexity, as claimed by~\citet{Steinert-Threlkeld2020Ease}, but rather the extent to which $\hat{P}$ is consistent with the convexity criterion.

\begin{figure}[t!]
    \centering
    \includegraphics[
    width=0.85\columnwidth,
    trim=0in 0.3in 0in 0in,    
    ]{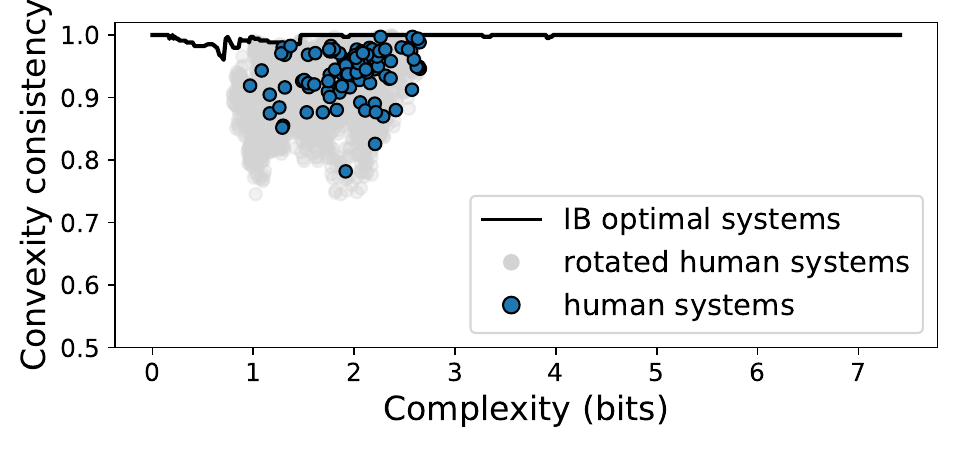}
    \caption{
    Convexity consistency (\Cref{eq:convexity_consistency}) of attested color naming systems (blue), rotated variants (gray) and optimal IB systems (black curve), as a function of their IB-complexity.
    }
    \label{fig:convexity_consistency}
\end{figure}

\Cref{fig:convexity_consistency} shows that indeed, as reported in earlier studies, the WCS systems and the IB-optimal ones are highly convex. However, our analysis also reveals two additional observations that challenge G{\"a}rdenfors' conjecture. First, while the IB-optimal systems are mostly convex, they occasionally exhibit convexity violations (where the black curve in \Cref{fig:convexity_consistency} dips below 1). While approximate convexity is expected when dealing with empirical data, for the optimally-derived systems that seems less expected if convexity is truly instrumental for efficiency. Second, when comparing the actual WCS systems with hypothetical variants obtained by rotations~\citep{Regier2007Color}, i.e., rotating each system along the hue dimension of the WCS color grid (\Cref{fig:stagesetting}a), we observe that also the rotated variants are highly convex. This suggests that convexity might be too permissive, as it is unclear whether it can explain why the actual WCS systems are attested while their rotated counterparts are not.

These two observations raise several important questions:
how are convexity and efficiency related to each other? Is one a necessary consequence of the other, or are they distinct in the sense that each can occur without the other? Does combining both properties offer substantially more explanatory power than characterizing semantic systems using only one? In what follows, we address these open questions by combining analytical results and further statistical data analysis.

\section{Convexity and efficiency are distinct: Neither is necessary for the other}

Our main analytical result is that convexity and efficiency do not entail each other. Mathematically, this amounts to showing that convex systems can be highly inefficient and that optimally efficient systems can be non-convex. In other words, the two properties are distinct from each other in the sense that neither is necessary for the other. This does not mean that efficiency and convexity cannot co-occur; as we have seen, a system can be both. However, it does imply that convexity does not facilitate efficiency, disconfirming G{\"a}rdenfors' conjecture. More crucially, it implies that although convexity and efficiency are largely compatible in the case of color, they might be incompatible in other domains. The full proofs are provided in \apndx~\ref{app:theoretical_results}. Here, we provide the main intuitions leading to the proofs and demonstrate their implications. 

To prove that efficiency does not entail convexity we construct a synthetic counter-example in which some optimal IB systems are highly non-convex (\apndx~\ref{app:theoretical_results}, Theorem \ref{app:thm1}). One intuitive interpretation of that synthetic example is the following scenario: suppose you wish to communicate to a friend whether they should walk straight (either forward or backward) or turn (either left or right). These concepts of `straight' and `turn' induce non-convex categories that could still be optimally efficient in the IB framework. This suggests that the cases in which the optimal IB color category systems are not perfectly convex are not meaningless but rather reflect cases where violating convexity supports greater efficiency. 

To show that convexity does not entail efficiency, we show that for any given convex system $P$ there exists an environment in which it is highly non-efficient (\apndx~\ref{app:theoretical_results}, Theorem \ref{app:thm2}). The construction of these environments is based on the communicative need distribution $p(u_t)$, which is a key component in efficiency-based theories of categorization~\citep{gibson2017, Kemp2018Semantic,zaslavsky2019need} but is not accounted for by the convexity theory. Because the hypothetical environments constructed in the proof are less intuitive, we demonstrate the consequence of this result by showing how to systematically construct highly inefficient yet perfectly convex color category systems.

To this end, we generated a large sample of Voronoi partitions, which are necessarily convex, using two distinct procedures across $k \in \{3, \dots, 17\}$ categories, matching the range observed in the WCS.  
First, we used a greedy swap-based search algorithm (\apndx~\ref{app:algorithms}, Algorithm \ref{app:alg1}), which begins by randomly sampling $k$ exemplars from the WCS grid (\Cref{fig:stagesetting}a) and constructs a Voronoi partition by mapping each color to its nearest exemplar in CIELAB space. It then greedily minimizes the IB accuracy term by iteratively replacing exemplars with other colors from the WCS grid that yield lower accuracy. We repeated this process across twenty random seeds for each $k$, recording all systems generated during these trajectories.  Second, we implemented a deterministic agglomerative clustering algorithm (\apndx~\ref{app:algorithms}, Algorithm \ref{app:alg2}): starting from a system where every WCS chip is its own category centroid ($k=330$), we iteratively merged pairs of categories whose mean CIELAB coordinate resulted in the lowest accuracy system of size $k-1$. By pooling the states from these procedures within our target $k$ range, we produced a sample of $N=971,525$ perfectly convex systems. For each system, we then measured its IB complexity, accuracy, communicative cost, and devaition from optimality. 

\Cref{fig:inefficient}a shows that a large number of convex systems are highly inefficient with respect to the IB bound. While \citet{Koshevoy2025Convexity} reported that systems approaching the IB bound also tend to be more convex, their sampling technique ignored convex systems lying far from the bound, thus yielding seemingly inconsistent results. Furthermore, \Cref{fig:inefficient}b shows that our results also extend to another closely related notion of efficiency, which replaces the IB complexity term by the number of major categories~\citep[RKK;][]{Regier2015Word}. That is, perfectly convex systems can be highly inefficient also with respect to the RKK color naming model. Figures \ref{fig:inefficient}e-g show that convex systems that are highly inefficient are also highly unnatural (see interactive versions of these figures and more exmples in our \SMWebApp).

\section{Efficiency explains convexity and beyond}

So far we have shown that efficiency and convexity are distinct properties and that in the case of color naming, while convexity appears to be too permissive, efficiency  explains much of the empirically observed structure in the WCS data, including convexity. This raises the question of which property may provide a more comprehensive empirical characterization, and to what extent we may gain explanatory power by considering both. To address these questions, we next explore how convexity and efficiency differ in their discrimination of actual color naming systems from hypothetical ones. For hypothetical systems, we consider the $39$ rotations (as in \Cref{fig:convexity_consistency}) of each attested system $P_l$, and denote these variants by $P_{l,r}$. If efficiency (\Cref{eq:epsilon}) and convexity (\Cref{eq:convexity_consistency}) characterize human category systems, then we would expect that attested systems would have an advantage over their hypothetical counterparts. That is, we would expect that they would be more efficient, i.e., $\varepsilon[P_{l,r}] - \varepsilon[P_{l}] > 0$ and more convex, i.e., $\mathrm{Conv}[\hat{P}_{l}] - \mathrm{Conv}[\hat{P}_{l,r}] > 0$.\footnote{For convexity, we consider the hard partitioning of a system, $\hat{P}_l$, and for efficiency, the language $l$'s full probabilistic structure.}

\begin{figure}[t!]
    \centering
    \includegraphics[
    width=0.95\columnwidth,
    % trim = <left> <bottom> <right> <top>
    trim=0in 11.5in 0in 0in,
    clip    
    ]{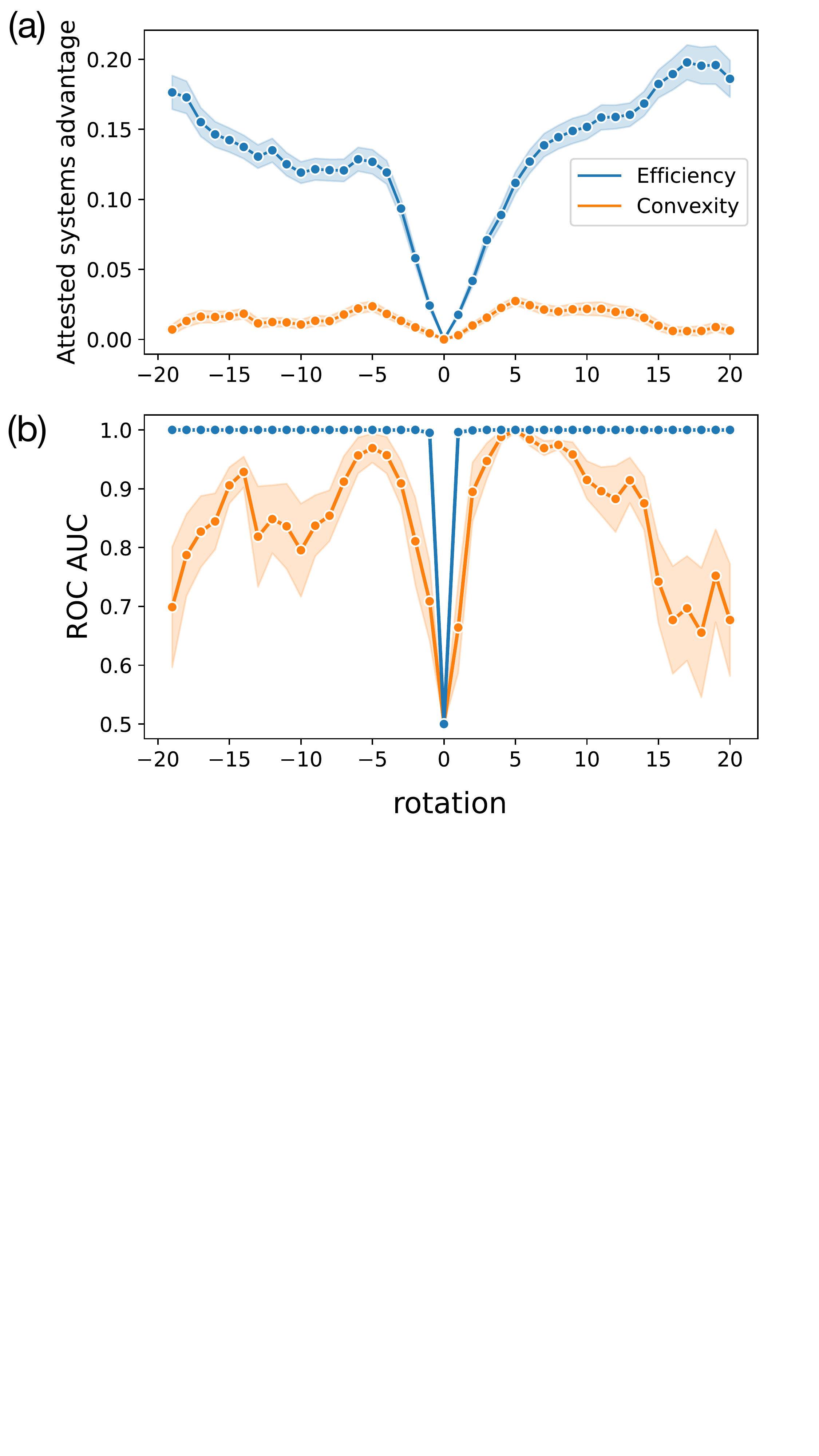}
    \caption{ \textbf{(a)} Advantage in efficiency (blue) and convexity consistency (orange) scores for rotations of attested systems along the hue dimension of the WCS grid. Shaded regions denote $95\%$ CIs. \textbf{(b)} Mean ROC AUC scores for classifiers trained on efficiency advantage and convexity advantage, evaluated over $5$-fold CV. 
    }
    \label{fig:rotations}
\end{figure}

\Cref{fig:rotations}a shows that efficiency yields much higher attested systems advantage than convexity. Only $13\%$ of attested systems exhibit a convexity advantage compared to all of their rotated variants, while $93\%$ show an efficiency advantage. This difference is also borne out in a discrimination task. We trained binary logistic regression classifiers to discriminate attested systems from their rotated variants using convexity and efficiency as features. That is, the classifiers observe either $\Delta\varepsilon[P_1, P_2] = \varepsilon[P_1]-\varepsilon[P_2]$ or $\Delta c = \mathrm{Conv}[P_1] - \mathrm{Conv}[P_2]$, and must predict which system is attested, $P_1$ or $P_2$. \Cref{fig:rotations}b shows the mean area under the receiver operating characteristic curve (ROC AUC)  over $5$-fold cross-validation (CV). While the efficiency-based classifier is at ceiling, the convexity-based classifier performs substantially worse. 

To test whether convexity may add explanatory power on top of efficiency, we performed a nested model comparison, comparing the efficiency-based and convexity-based classifiers to one that combines both features. For this analysis we report results over the full dataset, although similar results also hold for 5-fold CV. \Cref{tab:model_comparison} shows that the loss in log-likelihood when removing efficiency is over an order of magnitude greater than when removing convexity. In other words, efficiency explains the structure of attested systems, while convexity provides negligible incremental predictive power. Taken together with our findings in the previous section, these quantitative results suggest that a purely efficiency-based account may be sufficient to characterize human color naming without convexity as a necessary constraint~\citep[cf.,][]{Douven2020What}.

\begin{table}[t!]
    \centering
    \scriptsize
    \begin{tabular}{l cc c cc}
        \toprule
         & \multicolumn{2}{c}{Standardized Coefficients} & & \multicolumn{2}{c}{LRT (vs. full)} \\
        \cmidrule(lr){2-3} \cmidrule(lr){5-6}
        Model & $\Delta\varepsilon$ & $\Delta c$ & LL & $\chi^2(1)$ & $p$ \\
        \midrule
        \textbf{Full} ($\Delta\varepsilon + \Delta c$) & $12.09$ & $1.65$  & $-287.5$ & --- & --- \\
        \textbf{Efficiency} ($\Delta\varepsilon$) & $13.30$ & --- & $-309.0$ & $42.9$ & $<0.001$ \\
        \textbf{Convexity} ($\Delta c$) & --- & $2.27$  & $-4010.2$ & $7445.4$& $<0.001$ \\
        \bottomrule
    \end{tabular}
    \caption{Descriptive evaluation of model features via likelihood ratio tests (LRT) in discriminating attested systems from variants. $\Delta\varepsilon$ and $\Delta c$ are the differences in efficiency and degree of convexity, respectively, between the attested and hypothetical systems.
    }
    \label{tab:model_comparison}
\end{table}

\subsection{Phenomena beyond convexity}

While efficiency and convexity coincide in the case of color naming, it remains unclear whether this generalizes to other domains, especially those that pose a challenge to convexity but have been addressed in terms of efficiency. For example, in kinship, Northern Paiute has reciprocal grandparent-grandchild terms~\citep{Kroeber1917Kinship}. These terms will occupy non-convex regions of conceptual space, whether defined on the family tree or the feature-based geometry sketched in \citet{gardenfors2004knowledge}. In animal taxonomy, it has been documented that some languages distinguish \textit{fish} from \textit{all other life forms}~\citep{brown1984language}, but many life forms intuitively may lie on `opposite' sides of fish (e.g., sea and land mammals). Reconciling these phenomena with convexity seems challenging, while efficiency-based approaches account for these domains
~\citep{Kemp2012Kinship,Zaslavsky2019Semantic}. 

Apart from challenging domains, there are key phenomena that seem beyond the reach of convexity but have been accounted for in terms of efficiency.
An important example is how categories may adapt to specific environments. In IB and other efficiency-based approaches~\citep[e.g.,][]{Kemp2018Semantic,gibson2017}, a communicative need distribution, which reflects social and environmental communicative pressures, plays a central role. It was also shown that communicative efficiency can explain why languages in cold regions are less likely to group \textit{ice} and \textit{snow} in the same category~\citep{Regier2016Languages}. Convexity, however, does not incorporate any notion of communicative need, nor other social or environmental needs, and therefore cannot explain how semantic systems may adapt to different environments. Another important example is language evolution. While the IB framework has generated precise testable predictions for language evolution by allowing the tradeoff parameter $\beta$ to vary~\citep{Zaslavsky2018Efficient,Zaslavsky2022Evolution}, it is unclear how convexity might explain the evolution of categories over time. That is, simply maintaining convexity does not explain why or how categories evolve.

\section{Discussion}

Human semantic systems have been characterized as both efficient and convex. While these properties can empirically co-occur, their analytical relationship has remained unclear. This work has addressed this gap in the literature, combining analytical and empirical results. We showed that convexity and efficiency are distinct: a system can be perfectly convex yet highly inefficient, and optimally efficient systems can be non-convex. Furthermore, the fact that IB-optimal color naming systems are mostly convex can explain the empirical basis of convexity: as human category systems optimize for efficiency, they reflect the approximate convexity of optimal systems. This is consistent with our statistical analysis that showed that efficiency is a stronger predictor of attested color naming systems. We wish to highlight that \citet{gardenfors2000conceptual} himself acknowledged that convexity might not be a sufficient condition for natural categories, but claimed it may still be a necessary constraint~\citep{Douven2020What}. Our work suggests that this claim might not have empirical grounds, given that a purely efficiency-based account can explain empirically observed color categories, including their convexity, as well as other domains which are not necessarily convex and phenomena that seem beyond the reach of convexity, such as language evolution and adaptation to different environments.
Overall, our findings suggest that while convexity and efficiency may coincide, they are fundamentally distinct, with efficiency providing a more comprehensive account of semantic typology.
An important direction for future research is to extend our empirical comparisons of efficiency and convexity across more semantic domains.

\section{Acknowledgments} We thank Charles Kemp for valuable discussions and comments on the manuscript. 

\balance
\printbibliography

\newpage
\appendix
\clearpage
\nobalance
\setcounter{secnumdepth}{2} % enable numbering for Sections and Subsections
% \section{Appendix}

\section{Theoretical results: convexity and efficiency do not entail each other}
\label{app:theoretical_results}

\subsection{Preliminaries}
For a set $\mathcal{X} \subset \mathbb{R}^d$, denote by $\mathrm{co}(\mathcal{X})$ its convex hull, by $\Delta(\mathcal{X})$ the simplex over $\mathcal{X}$, and by $\Delta ( \mathcal{Y} )^{\mathcal{X}} $ the simplex of all conditional probabilities $p(y|x)$. 
Let $\cU \subset \mathbb{R}^d$ be a semantic domain defined over a $d$-dimensional space of real-valued features, and $\mathcal{W}_k = \{ 1, \dots, k \}$ be a set of categories (or labels). 

\definition{A hard $k$-category system over $\mathcal{U}$ is defined by a mapping $P: \mathcal{U} \rightarrow \mathcal{W}_k $.}

\definition{ A soft $k$-category system over $\mathcal{U}$ is defined by a probabilistic mapping $q(w|u) \in \Delta ( \mathcal{W}_k )^{ \mathcal{U} } $. }

\definition{For a given $\beta\ge0$, prior $p(u_t)$, and speaker's representations $m_t(u)$ for $u_t\in\cU$, an IB-optimal category system is $q_{\beta}(w|u) =\argmin_q \F_{\beta}[q]$.
}

\definition{ For a soft category system $q$, we define its corresponding hard system by $Q(u) = \arg\max_{w} q(w | u)$, assuming for simplicity that the $\arg\max$ value is unique. Note that $Q$ is a $k'$ system with $k' \leq k$. }

\definition{ (Category extension). Given a system $P$ and a category $w \in \mathcal{W}$, we define the category extension of $w$ by  
\begin{equation}
    C_P(w) = \{ u: P(u) = w \}.
\end{equation}
For a soft system $q$, we take $C_Q(w)$ to be its category extension. 
}

\subsection{Proofs}
\label{app:proofs}

Consider a domain that is defined by varying angles, which can be represented either as a circle or as the line segment $[0, 2\pi]$. Intuitively, this could capture a hue circle, or a walking direction (as in our example in the main text), or a simplified version of the so-called ``Shepard circles'' domain \citep{Shepard1964Attention} where only the angle of the stimulus varies and the radius is fixed. There are infinitely many ways to partition this domain into $k$ convex regions. From a convexity perspective, all of these partitions are equally good and ideal. From an efficiency perspective, however, they are not all equally good. It turns out, that there are also efficient ways to partition this space which are not convex. The following two theorems show this. More formally, they show that convexity and efficiency do not entail each other. 

% \subsection{Efficiency does not entail convexity}
\theorem[Efficiency does not entail convexity]{Optimal IB systems are not necessarily convex. That is, there exists a
non-convex system that is IB-optimal. }
\label{app:thm1}

\begin{figure}[H]
    \centering
    \includegraphics[
    width=.9\columnwidth,
    trim = 0 0 0 0,
    ]{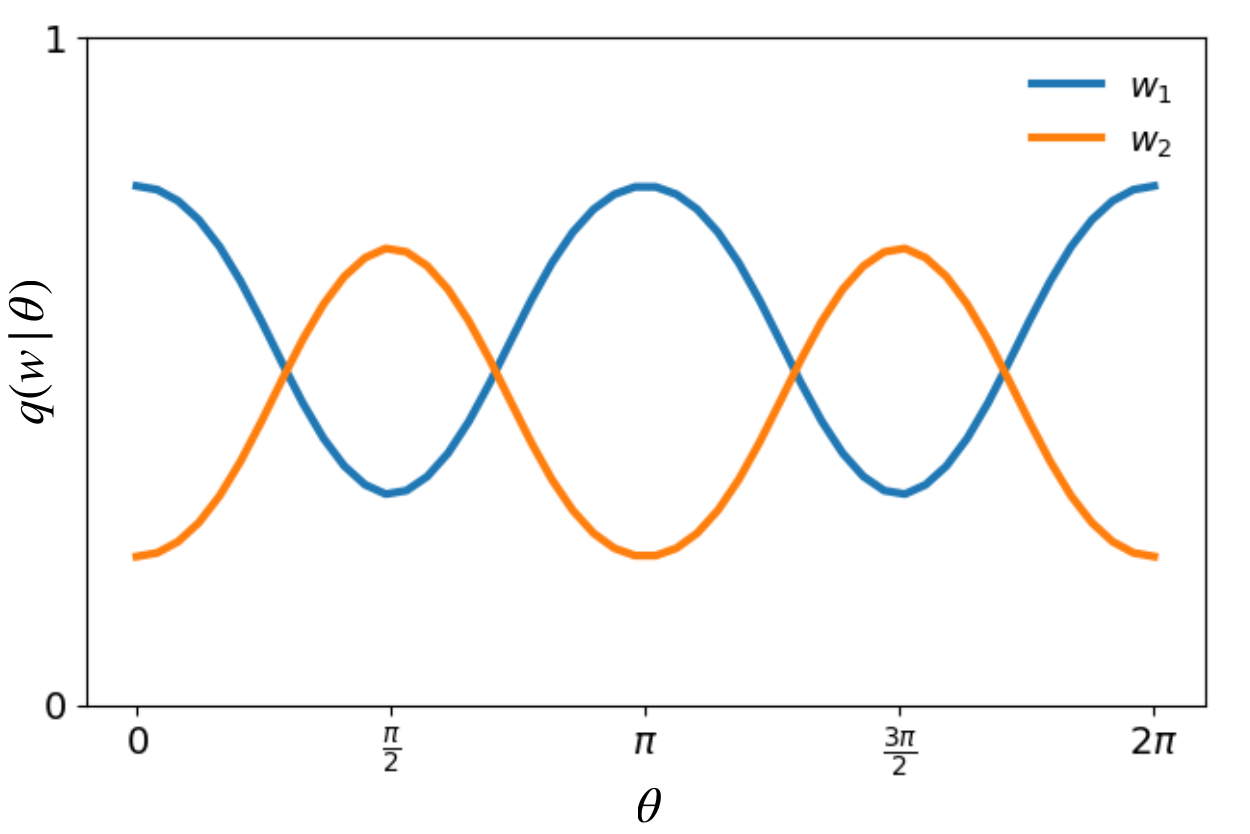}
    \caption{Example of a non-convex IB optimal system ($\beta \approx 9.11$) with $k=2$ categories. The categories are periodic with crossing boundaries.
    }
    \label{fig:numerical_example}
\end{figure}

\begin{proof}
    Consider the angle domain described above, i.e., $\cU=[0,2\pi]$ with a uniform prior over this interval and similarity-based speaker representations, $m_t(u)\propto\exp(\mathrm{Sim}(u,u_t))$, where $\mathrm{Sim}(u, u') = |\cos(u - u')|$. Intuitively, this representation of the space reflects indifference to the direction of the angle.
    Let $m_{\theta}(u)$ be the representation of $u_t=\theta$. It is easy to verify that $m_0 \equiv m_{\pi}$ and $m_{\frac{\pi}{2}}\equiv m_{\frac{3\pi}{2}}$.
    
    The IB-optimal systems must have the following form~\citep{Tishby1999Information}:
    \eqarr{
        q(w|m_t) &= \frac{q(w)}{Z_t}\exp(-\beta D[m_t(u)\|m_w(u)]).
    }
    Therefore, $q(w|m_0) = q(w|m_{\pi})$ and $q(w|m_{\frac{\pi}{2}}) = q(w|m_{\frac{3\pi}{2}})$ for all $w$. Suppose there is an IB system for which $Q(u)$ is a 2-category system where both categories, $w_1$ and $w_2$, have at least one angle $u_i$ for which $w_i = Q(u_i)$, then this system cannot be convex. To see this, assume for contradiction that it is convex. Let $w_1 = Q(0)=Q(\pi)=Q(2\pi)$. If $w_2 = Q(\frac{\pi}{2})$ then the system is non-convex, which would be a contradiction. Therefore, by our assumption, it must hold that $w_1=Q(\frac{\pi}{2}) = Q(\frac{3\pi}{2})$. However, $u_2$ is necessarily in one of the intervals $[0,\frac{\pi}{2}]$, $[\frac{\pi}{2}, \pi]$, $[\pi, \frac{3\pi}{2}]$, $[\frac{3\pi}{2}, 2\pi]$, which contradicts the convexity assumption. To conclude the proof, we show numerically that such an IB system exists (\Cref{fig:numerical_example}).
\end{proof}

Our second theorem shows that convexity does not entail efficiency by showing that for any convex $k$-category system there is an environment where that system is highly inefficient. It also shows that for the same environment, there is a different convex system that is maximally efficient, thus the problem is not with the environment itself. This proves that convexity cannot explain adaptation to different environments and cannot distinguish between maximally efficient and maximally inefficient systems.

\theorem[Convexity does not entail efficiency]{For any convex $k$-category system $P$ there exists a world in which $P$ is non-informative,
while it is possible to attain maximal informativeness, i.e., $\log k$, with another convex $k$-category system $Q$.} 
\label{app:thm2}

\begin{proof}
    Consider the same angle domain as before, but here we will assume speaker certainty, i.e., $m_t(u)=\delta_{u_t,u}$. For a given convex system $P$, we will (1) construct a prior for which $P$ is non-informative and (2) show that there is another system which attains maximal informativeness. To this end, let $p(u_t)$ be a uniform prior over a sub-segment $C_A \subseteq C_P(w^*)$ of length $A$, for $w^* \in \mathcal{W}_k$. Note that $C_P(w)$ is a line segment because $P$ is convex and our domain is one-dimensional. Denote the length of a segment $I$ by $|I|$, i.e., $|C_A|=A$. 
    In this case, the informativeness, or accuracy, of any convex system $Q$ is reduced to
    \begin{align}
        I_Q(W;U) &= \sum_w q(w) \int_{\theta \in C_A} q(\theta | w) \log \frac{q(\theta | w)}{p(\theta)} d\theta\\
        &= \sum_w \frac{ |C_Q(w) \cap C_A| }{ A } \log \frac{A}{ | C_Q(w) \cap C_A  | }
    \end{align}
    because 
    \eqarr{
        q(w) &= \frac{|C_Q(w) \cap C_A|}{A}\\
        q(u|w) &= \frac{q(w|u)p(u)}{q(w)} = \frac{\delta_{Q(u),w}}{|C_Q(w) \cap C_A|}\,.
    }
    For $P$  we get that $I_P(W;U)=0$ because $|C_P(w^*) \cap C_A| = A$ and for all $w\neq w^*$, $|C_P(w^*) \cap C_A| = 0$. This means that $P$ has minimal accuracy. Maximal accuracy is achieved when $Q$ partitions $C_A$ into $k$ equal segments, i.e. $| C_Q(w) \cap C_A | = \frac{A}{k}$ for all $w\in\W_k$, in which case $I_Q(W;U)=\log k$. 
\end{proof}

Strictly speaking, the proof above constructs a prior such that the IB-complexity of $P$ is also 0, which means that in IB-terms, $P$ is trivially inefficient but also trivially efficient. However, it is maximally inefficient with respect to a closely related notion of efficiency, the RKK formulation~\citep{Regier2015Word}, where complexity if measured by the number of categories $k$. Finally, we note that our work also presents an algorithmic method of finding convex systems that are highly inefficient with non-trivial IB-complexities (see main text), which complements this proof in showing that convexity does not entail efficiency.
\vfill

\section{Algorithms for finding inefficient convex category systems}
\label{app:algorithms}

\begin{algorithm}[H]
\caption{ Exemplar search via greedy swapping }
\begin{algorithmic}[1]
\Require $X = \{x_1, \dots, x_{330}\} \subset \mathbb{R}^3$ (CIELAB), subset size $k$, max iterations $M$, prior $p(M)$, meanings $p(U|M)$.
\Ensure Final exemplar set $E$, trace of systems $T$, convergence flag $C$.

\For{$k = 3$ \textbf{to} $18$}
    \For{$s = 1$ \textbf{to} $20$}
    % \State $(E, T, C) \gets \text{\textsc{GreedySwapSearch}}($
        % \Statex \hspace{2.5em} $X, k, M, \text{seed})$
        \State $(E, T, C) \gets \text{\textsc{GreedySwapSearch}}(X, k, M, s)$
        \State Store results
    \EndFor
\EndFor

\Statex
\Function{GreedySwapSearch}{$X, k, M, s$}
    \State Set random state using seed $s$
    \State $E \gets \text{random subset of } X \text{ with size } k$
    \State $T \gets \{E\}$, $C \gets \text{False}$, $t \gets 0$

    \While{$C = \text{False}$ \textbf{and} $t < M$}
        \State $C \gets \text{True}$ \Comment{Assume convergence first}
        \State $t \gets t + 1$
        \State $E_{\mathrm{shuffled}} \gets \text{shuffle}(E)$
        
        \ForAll{$x_i \in E_{\mathrm{shuffled}}$}
            \State $X_{\mathrm{rem}} \gets X \setminus E$
            \State $dA_{\mathrm{max}} \gets -\infty$
            \State $x_{j^*} \gets \text{null}$
            
            \State \Comment{Find best chip to swap with $x_i$}
            \ForAll{$x_j \in X_{\mathrm{rem}}$}
                \State $E_{\mathrm{cand}} \gets (E \setminus \{x_i\}) \cup \{x_j\}$
                \State $dA \gets \text{Acc}(E_{\mathrm{cand}}) - \text{Acc}(E)$
                \If{$dA > dA_{\mathrm{max}}$}
                    \State $dA_{\mathrm{max}} \gets dA$
                    \State $x_{j^*} \gets x_j$
                \EndIf
            \EndFor
            
            \If{$dA_{\mathrm{max}} > 0$}
                \State $E \gets (E \setminus \{x_i\}) \cup \{x_{j^*}\}$
                \State $T \gets T \cup \{E\}$ \Comment{Record trace}
                \State $C \gets \text{False}$
            \EndIf
        \EndFor
    \EndWhile
    \State \Return $E, T, C$
\EndFunction

\\
\Function{Acc}{$E$} \Comment{Helper function for IB accuracy of a convex partition}
    \State \Comment{1. Create Voronoi partition $p(W|M)$ based on Euclidean distance in CIELAB}
    \For{\textbf{each} $m \in X$}
        \State $w^* \gets \arg\min_{e \in E} \|m - e\|_2$ 
        \State Set $p(w^*|m) \gets 1$ and $p(w \neq w^*|m) \gets 0$
    \EndFor
    
    \State $\mathrm{acc} \gets I(W;U) \text{ given } p(W|M), p(M), \text{ and } p(U|M)$
    % \State \Return $I(W;U)$ given $p(W|M), p(M), \text{ and } p(U|M)$
    \State \Return acc
\EndFunction

\end{algorithmic}
\label{app:alg1}
\end{algorithm}

\begin{algorithm}[H]
\caption{Deterministic agglomerative merging}
\begin{algorithmic}[1]
\Require Dataset $X = \{x_1, \dots, x_{330}\} \subset \mathbb{R}^3$,  prior $p(M)$, meanings $p(U|M)$.
\Ensure Trace of systems $\mathcal{T} = \{C_{330}, \dots, C_3\}$.
\Statex
\State $C \gets X$ \Comment{Initial centroids are the 330 CIELAB points}
\State $\mathcal{T}[330] \gets C$ 

\For{$k = 329$ \textbf{down to} $3$}
    \State $min\_acc \gets \infty$
    \State $best\_C \gets \emptyset$
    \State $n \gets |C|$
    
    \State \Comment{Loop over all pairs in the current centroid set}
    \For{$i = 1$ \textbf{to} $n$}
        \For{$j = i + 1$ \textbf{to} $n$}
            \State $c_{new} \gets \frac{c_i + c_j}{2}$ \Comment{Merge via CIELAB average}    
            \State $C' \gets (C \setminus \{c_i, c_j\}) \cup \{c_{new}\}$
            
            \If{\textsc{CountCategories}($C', X$) $= k$}
                \State $curr\_acc \gets \textsc{Acc}(C', p(M), p(U|M))$
                \If{$curr\_acc < min\_acc$}
                    \State $min\_acc \gets curr\_acc$
                    \State $best\_C \gets C'$
                \EndIf
            \EndIf
        \EndFor
    \EndFor
    \State $C \gets best\_C$
    \State $\mathcal{T}[k] \gets C$ \Comment{Store the min acc. system for $k$}
\EndFor

\State \Return $\mathcal{T}$

\Statex
\Function{CountCategories}{$C, X$}
    \State $V \gets \emptyset$ \Comment{Set of centroids that correspond to chips}
    \For{\textbf{each} $u \in X$}
        \State $c^* \gets \arg\min_{c \in C} \|u - c\|_2^2$
        \State $V \gets V \cup \{c^*\}$
    \EndFor
    \State \Return $|V|$
\EndFunction
\end{algorithmic}
\label{app:alg2}
\end{algorithm}

\end{document}

%% file: header_modified.tex
\usepackage[dvipsnames, svgnames, x11names]{xcolor}
\usepackage{algorithm}
\usepackage{amsfonts}
\usepackage{tikz}
\usepackage{amsmath}
\usepackage{amsthm}
\usepackage[labelfont={bf,small},labelsep=period,textfont={small}]{caption}
\usepackage{graphicx}

\usepackage[bookmarks=true, colorlinks, breaklinks]{hyperref}
\AtBeginDocument{\hypersetup{
 linkcolor={NavyBlue},
 citecolor={NavyBlue},
 urlcolor={NavyBlue},
 }}

\usepackage[hang,flushmargin]{footmisc}

\DeclareMathOperator*{\argmin}{argmin}
\DeclareMathOperator*{\argmax}{argmax}

 \newcommand{\reals}{\mathbb{R}}

\newcommand{\cU}{\mathcal{U}}
\newcommand{\F}{\mathcal{F}}

 \newcommand{\W}{\mathcal{W}}

\newcommand{\eqn}[1]{\begin{equation}#1\end{equation}}
\newcommand{\eqarr}[1]{\begin{align}#1\end{align}}